\documentclass[a4paper, 10pt, conference]{ieeeconf}      
\usepackage{FG2024}

\FGfinalcopy 

\IEEEoverridecommandlockouts                              
\usepackage{graphicx} 
\usepackage{multirow}

\usepackage{amsmath}
\usepackage{makecell} 
\usepackage{balance} 
\usepackage{hyperref}

\usepackage{fancyhdr}

\def\FGPaperID{162} 

\title{\LARGE \bf
Distilling Privileged Multimodal Information for Expression \\ Recognition using Optimal Transport}

\author{\parbox{16cm}{\centering
    {\large Muhammad Haseeb Aslam$^1$ , Muhammad Osama Zeeshan$^1$, Soufiane Belharbi$^1$,\\ Marco Pedersoli$^1$, Alessandro Lameiras Koerich$^1$, Simon Bacon$^2$ and Eric Granger$^1$}\\
    {\normalsize
    $^1$ LIVIA, Dept. of Systems Engineering, ETS Montreal, Canada\\
    $^2$ Dept. of Health, Kinesiology $\&$ Applied Physiology, Concordia University, Montreal, Canada}
    {\tt \small
     muhammad-haseeb.aslam.1@ens.etsmtl.ca \hspace{1cm}   eric.granger@etsmtl.ca
    }
}
}

\begin{document}

\ifFGfinal
\thispagestyle{empty}
\pagestyle{empty}
\else
\author{Anonymous FG2024 submission\\ Paper ID \FGPaperID \\}
\pagestyle{plain}
\fi
\maketitle
\thispagestyle{fancy}

\begin{abstract}
Deep learning models for multimodal expression recognition have reached remarkable performance in controlled laboratory environments because of their ability to learn complementary and redundant semantic information. However, these models struggle in the wild, mainly because of the unavailability and quality of modalities used for training. In practice, only a subset of the training-time modalities may be available at test time. Learning with privileged information enables models to exploit data from additional modalities that are only available during training.  
State-of-the-art knowledge distillation (KD) methods have been proposed to distill information from multiple teacher models (each trained on a modality) to a common student model. These privileged KD methods typically utilize point-to-point matching, yet have no explicit mechanism to capture the structural information in the teacher representation space formed by introducing the privileged modality. 
We argue that encoding this same structure in the student space may lead to enhanced student performance. This paper introduces a new structural KD mechanism based on optimal transport (OT), where entropy-regularized OT distills the structural dark knowledge. Our privileged KD with OT (PKDOT) method captures the local structures in the multimodal teacher representation by calculating a cosine similarity matrix and selecting the top-k anchors to allow for sparse OT solutions, resulting in a more stable distillation process. 
Experiments\footnote[1]{Code: \url{https://github.com/haseebaslam95/PKDOT}} were performed on two challenging problems -- pain estimation on the Biovid dataset (ordinal classification) and arousal-valance prediction on the Affwild2 dataset (regression). Results show that our proposed method can outperform state-of-the-art privileged KD methods on these problems. The diversity among modalities and fusion architectures indicates that PKDOT is modality- and model-agnostic. 
\end{abstract}

\section{Introduction}

Multimodal expression recognition (MER) aims to capture and mimic the human-like nature of emotions \cite{Shon2018}. The complementarity and redundancy of the multiple modalities typically allow all these MER systems to outperform their unimodal counterparts \cite{TensorFN, Praveen2022AudioVisualFF}. Using multiple modalities works well in a lab-controlled environment where all the modalities are available during testing and inference. However, some modalities are difficult or expensive to obtain in real-world deployment. Various strategies, like cross-attention \cite{rajasekhar} that weigh the modalities dynamically, have been proposed in the literature to mitigate these issues. However, even those methods are not effective where some modalities are entirely missing. Physiological data like electroencephalogram (EEG) \cite{eeg} electrocardiogram (ECG) and electromyography (EMG) \cite{Phan_biovid_phy} are widely studied in affective computing, and various MER systems have been proposed to recognize expressions. For instance, the physiological signals have been shown to outperform the visual modality, especially in tasks like pain estimation \cite{thiam_biovid}. However, these signals are not always available in some practical contexts. Since they require specialized equipment and limit the mobility of the subjects, in such cases, most methods will typically rely on prevalent modalities (available at both design time and development) resulting in a less effective system. Using additional information (privileged modalities) only at the training time, may however, enhance system performance.

\begin{figure}[!t]
 \centering
  \includegraphics[width=1\linewidth]{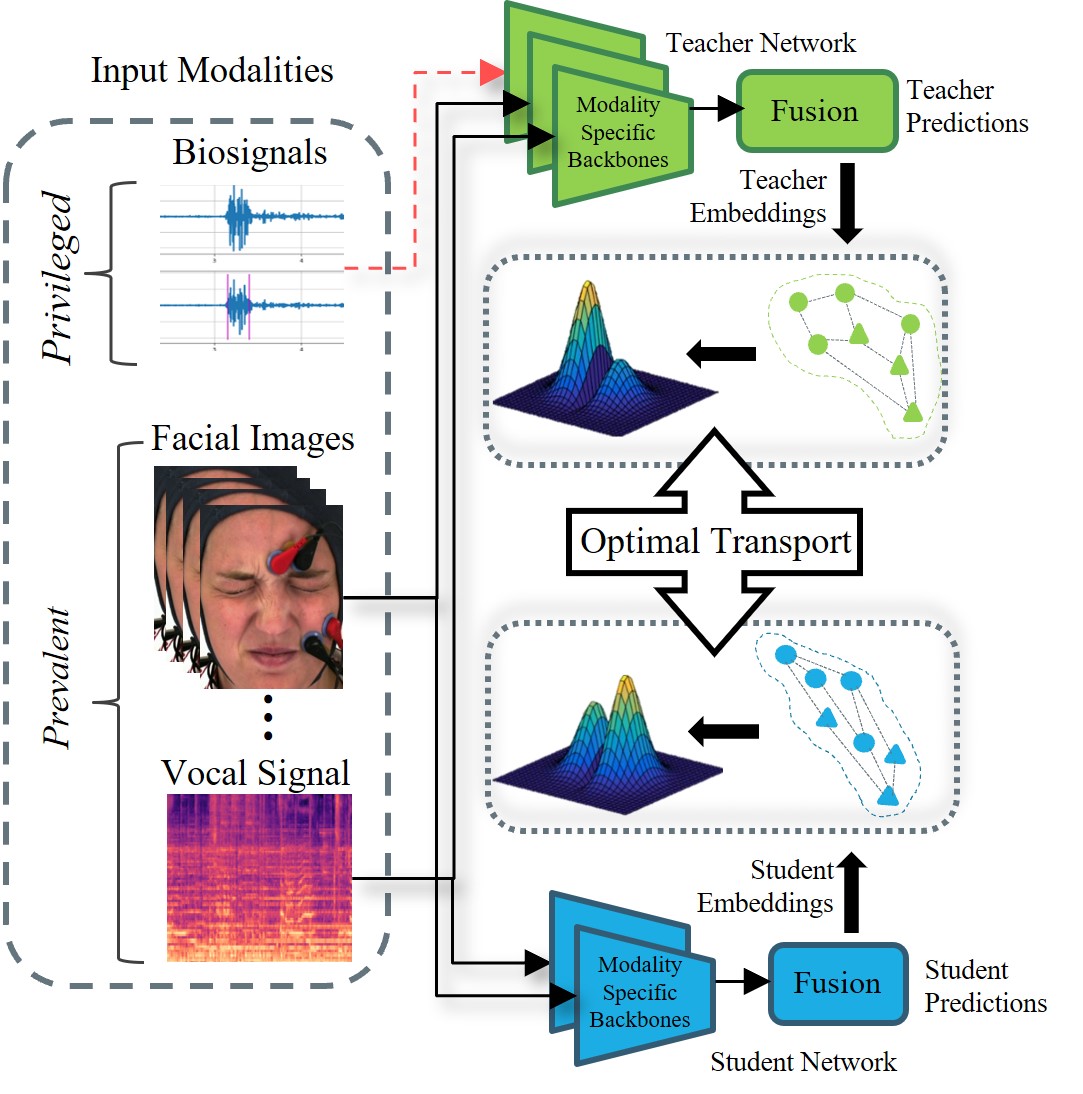}
  \caption{Overview of the proposed Privileged Knowledge Distillation with Optimal Transport (PKDOT) method that captures the local structure in the multimodal teacher representation. Teacher backbones process multimodal input data that is privileged (red arrow) and prevalent (black arrows), while student backbones only process prevalent modalities}
  \label{fig:top-pkdot}
  \vspace{-15pt}
\end{figure}

Recently, the learning using privileged information (LUPI) paradigm has been introduced for affective computing \cite{pkd-makantasis}. Privileged information in machine learning (ML) is the information available to the model only at training time but not at inference time \cite{lupi}. LUPI methods have been shown to increase performance for systems using only the prevalent modalities (available during train and inference). Traditionally utilizing teacher-student networks, the teacher network is an MER model trained with all the modalities, and this model is then used to distill the knowledge to an unimodal model that only inputs the prevalent modality \cite{pkd-makantasis,pkd-aslam}. One way to distill this information to the student is using the vanilla knowledge distillation (KD) method, which computes the Kullback–Leibler (KL) divergence between the softened logits in the teacher and student network. Other methods have proposed matching teacher and student representations in the feature space using Mean Squared Error (MSE) or cosine similarity. All of these methods use point-to-point matching to distill the knowledge, and are therefore, unable to capture the local structure formed in the teacher space. \textit{Local structure} refers to a more fine-grained knowledge representation, where the distance of each sample from all the other samples in the training batch is calculated. We argue that this structural information, formed in the teacher space through the interaction between the prevalent and privileged modality data, should be captured to mimic the performance in the student space. 

In this paper, a Privileged Knowledge Distillation with Optimal Transport (PKDOT) method is proposed that relies on a structural KD mechanism to effectively capture and distill the structural dark knowledge based on OT for the student (See Fig.~\ref{fig:top-pkdot}). Privileged KD based on a similarity matrix computes the similarity over a training batch rather than flattening and computing a single cosine similarity score. The cosine similarity matrix retains information about the pairwise relationships between all batch samples. An encoder-decoder transformation network (T-Net) is trained during the teacher training stage, and is employed to hallucinate the privileged modality features. Entropy-regularized OT is then used to distill the structural dark knowledge.

The main contributions of this work are as follows. (1) A PKDOT method is introduced that computes a cosine batch similarity matrix to distill knowledge from teacher to student networks using entropy-regularized OT on selected anchor points to introduce sparsity. T-Net is also introduced that hallucinates the privileged modality at test time. (2) Extensive experiments were performed on two challenging MER problems -- pain estimation on the Biovid dataset and arousal-valance prediction on the Affwild2 dataset. Results with different combinations of teacher and student modalities and fusion architectures indicate that the proposed PKDOT method allows for outperforming SOTA-privileged KD methods, and is modality- and model-agnostic.

\section{Related Work}

\subsection{Multimodal Expression Recognition (MER)}
MER is the understanding and modeling of emotions through jointly learning from different signals, including facial cues, audio, and text. Ngiam et al.~\cite{ngiam} proposed the seminal work in multimodal DL, where visual and audio modalities were encoded separately. Restricted Boltzmann Machines and autoencoders were then used to fuse this information. 
Tzirakis et al.~\cite{tzirakis} proposed an audio-visual fusion model, where a 1D convolutional neural network (CNN) was used to extract features from the audio modality, and a ResNet-50 CNN was used for visual feature extraction. The two feature vectors were concatenated and fed into a recurrent network for simultaneous temporal modeling and fusion. 
Rajasekhar et al.~\cite{rajasekhar} proposed a joint cross-attention mechanism for audio-visual fusion, where separated backbones were trained for audio and video modalities. The audio was transformed into spectrograms using the discrete Fourier transform (DFT) and fed to a ResNet-18 CNN, and an I3D CNN was used to extract features from the visual modality. Later, both the feature vectors were fed to a cross-attention module to calculate attention weights. The raw features were then multiplied with the attention weights to get attended features. This dynamically assigned weights to these modalities to overcome the noisy modality problem. The higher performance and robustness of the multimodal systems come at a cost, usually in terms of time and computational complexity. Recent efforts in minimizing the cost of multimodal systems have birthed concepts like attention bottleneck in transformers \cite{nagrani}. These methods still fall short in cases where some modalities are entirely missing. 

Pain estimation is another major affective computing problem with many publicly available datasets, with the recent addition of pain estimation in infants \cite{pain_inf} and animals \cite{pain_anim}. Many physiological and image-based approaches for pain detection are proposed. Werner et al.~\cite{werner_2017} developed a pain-specific feature set called facial activity descriptors.  
Dragomir et al.~\cite{dragomir} proposed a subject-independent deep learning approach with residual learning for pain estimation from facial images. Morabit et al.~\cite{morabit-biovida} proposed a data-efficient image transformer. Zhi et al.~\cite{zhi-biovid} proposed a sparse LSTM-based method to solve the vanishing gradient problem. Phan et al.~\cite{Phan_biovid_phy} proposed an attention-based method to process the EDA and ECG signals for pain estimation in the physiological domain. Multi-level context information was extracted using LSTM and attention network and then fused at the decision level for final pain prediction. Lu et al.~\cite{lu_biovid_phy} proposed a transformer and squeeze and excitation-based network to process multiscale EDA signals. Specifically for pain estimation, the physiological signal-based approaches have been shown to outperform the facial image-based approaches. Zhu et al.~\cite{Zhi2021Multimodal_biovid}, Kachele et al.~\cite{Kchele2015BioVid_mm}, and Werner et al.~\cite{wener-multimodal-2014} have proposed methods to combine visual and physiological modalities. The ultimate goal of this paper is to achieve performance that approaches multimodal systems using only the prevalent modality at test time. 

\subsection{Learning Using Privileged Information}
Vapnik and Vashist \cite{lupi} introduced the concept of learning using privileged information in ML. Additional information, only available at training time, was used to learn more discriminative information, outperforming the traditional ML paradigm of using the same information at training and testing. Many applications, like person re-identification \cite{pi-shang} and action recognition \cite{prid-pi}, have since employed this concept to enhance performance or increase robustness. Zhao et al.~\cite{zhao-pi} proposed a privileged KD mechanism for online action detection. Future frames from the video were used as privileged information only at training time since the model can access future frames during training; however, only historical frames are available in the real world. To reduce the teacher-student gap, only partial hidden features of the student model were updated through KL-divergence loss. The availability and quality of the modalities are a concern in the multimodal ML paradigm. In the case of RGB-D data, depth sensing is available for the training data, but in the wild, only RGB is commonly available. Audio and visual modalities can be partially missing due to user-initiated muting, transmission/recording errors, occlusion, etc. Similarly, in the affective computing domain, some modalities, e.g., ECG, EMG, and EDA, may be completely absent at test time.

\subsection{Knowledge Distillation and Optimal Transport}
The seminal work in knowledge distillation was proposed by Hinton et al.~\cite{Hinton2015} for model compression in a teacher-student learning paradigm. The accurate yet cumbersome teacher model is used as additional supervision to train a lightweight student model. The 'dark knowledge' comes from increasing the temperature in the teacher's softmax, making it less confident about one particular class. These softened logits encompass more information than one-hot encoding. Romero et al.~\cite{Romero2014} proposed hint learning, essentially distilling from the model's hidden layer instead of softened logits. Since then, several works for feature-based knowledge distillation have been proposed \cite{Kim2018ParaphrasingCN, heo-ft}. Vanilla KD methods only consider individual samples and solely rely on matching the output activations between teacher and student. Park et al.~\cite{Park2019RelationalKD} proposed RKD and argued that distilling the relational knowledge among samples can significantly improve the student's performance. Angle-wise and distance-wise loss were incorporated to penalize the structural differences between the teacher and student. Following that, this work also takes into account the local structures of the teacher model and effectively distills into the student model using OT.   

OT is a well-established mathematical framework for calculating the optimal cost of transforming one probability distribution into another \cite{Villani2008OptimalTO}. OT has been directly applied to ML problems where matching distributions are vital. Other solutions of distribution matching, like maximum mean discrepancy (MMD) and KL-divergence, suffer theoretical drawbacks \cite{feydi-iopsd}. MMD does not precisely capture the distance between distributions due to its sensitivity to sample size and outliers \cite{mmd_db_gretton}. KL-divergence has been shown to fail in cases where two distributions do not overlap, resulting in infinity \cite{kl-db-lohit}. Although an expensive solution, OT provides a stable metric for matching distributions. To overcome the statistical and computational limitations of the OT, Cuturi et al.~\cite{Cuturi2013SinkhornDL} proposed regularized OT. Many works have used OT of various applications, including model compression \cite{kl-db-lohit}, domain adaptation \cite{Luo_2023_CVPR_DA}, pedestrian detection \cite{Song_2023_CVPR-ped-det}, and neural architecture search \cite{Yang_2023_CVPR_NAS}. To the best of our knowledge, this is the first use of OT for privileged KD in the context of expression recognition. 
\begin{figure}[!t]
 \centering
  \includegraphics[width=0.95\linewidth]{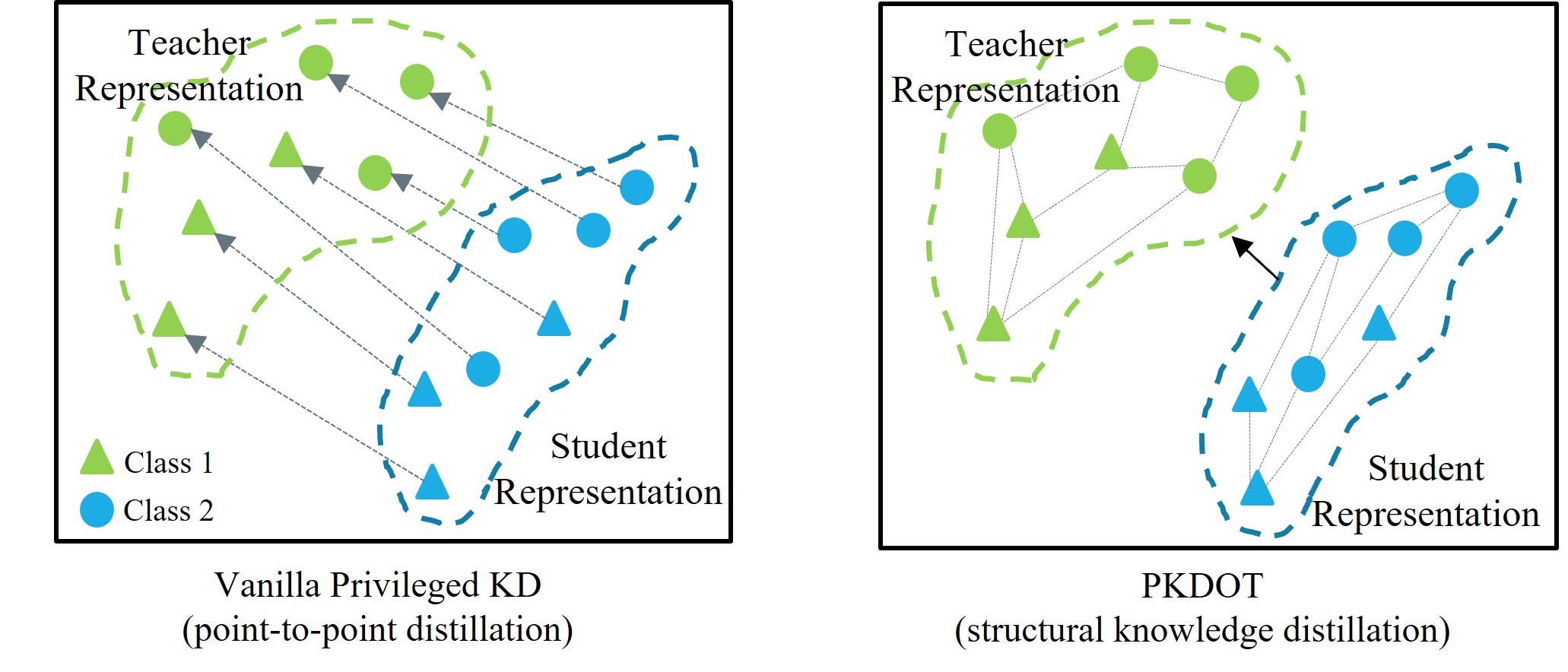}
  \caption{(Left) Conventional privileged KD computes point-to-point distance without considering local structure. (Right) The proposed PKDOT method captures the local structure and matches teacher and student representations by distilling the structural dark knowledge (adapted from \cite{Park2019RelationalKD}).}
  \label{fig:pkd-pkdot}
  \vspace{-10pt}
\end{figure}

\subsection{Privileged KD for Expression Recognition}
Recently, expression recognition has seen a rise in privileged KD methods as the trend has shifted from more in-lab scenarios to more in-the-wild scenarios. These works mainly include using the privileged information as an additional modality to train a superior multimodal model, which is then used to distill to a student model that does not have access to the privileged modality. Aslam et al.~\cite{pkd-aslam} proposed a model agnostic method for distilling privileged information (PI). Cosine similarity-based loss was added to the task loss with adaptive weighting to mitigate the negative transfer. Makantasis et al.~\cite{pkd-makantasis} proposed a similar valence and arousal prediction method using an MSE loss function and a categorical expression problem using a KL-divergence loss function. Liu et al.~\cite{Liu-pkd} also proposed a KD-based privileged DL mechanism in the physiological domain, where the teacher model was trained using the electroencephalogram (EEG) and galvanic skin response (GSR) signals. Later, this teacher model (EEG+GSR) was used to distill knowledge to the student model (GSR only) using KL-divergence. Although they enhance the student model performance, these methods cannot capture the structural knowledge in the multimodal teacher space. Capturing the local structures formed in the teacher network space by the introduction of privileged modalities should be distilled to the student network to enhance performance. Conventional KD methods rely on point-to-point matching. In this paper, we propose a structural KD mechanism using OT. Fig.~\ref{fig:pkd-pkdot} contrasts point-to-point vs. structural KD.   

\addtolength{\textheight}{-3cm}

\section{Proposed Approach}
\begin{figure*}[!t]
  \centering
   \includegraphics[width=.89\linewidth]{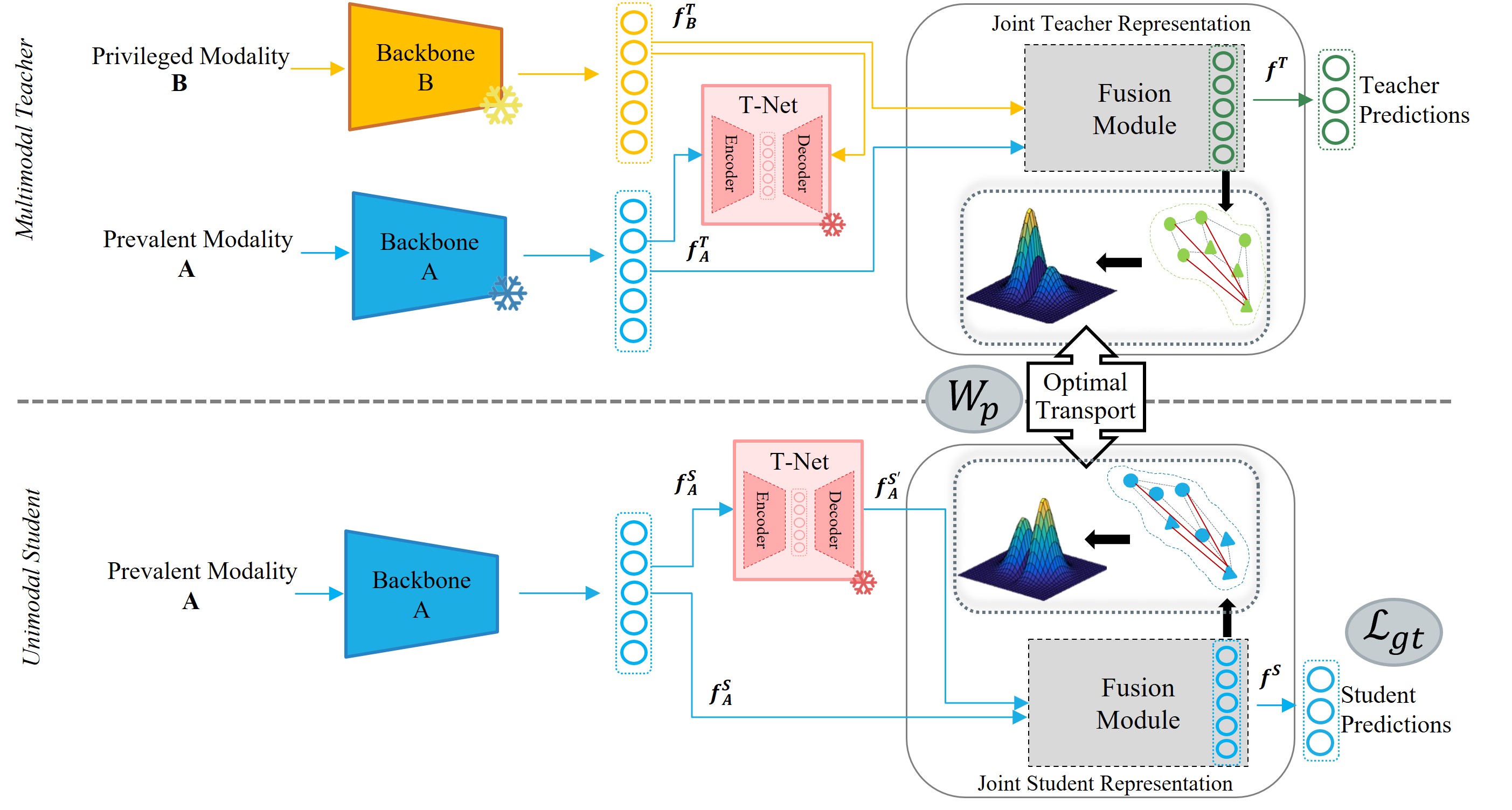}
   \caption{Illustration of the proposed PKDOT method with prevalent and privileged modality backbones and fusion. The teacher network (top) is trained on both prevalent and privileged modalities, while the student network (bottom) only inputs the prevalent modality. It hallucinates the features of the privileged modality and generates student embeddings in the multimodal space. Entropy-regularized OT is used to distill the structural dark knowledge.}
   \label{fig:pkdot_main}
   \vspace{-10pt}
\end{figure*}

The Privileged Knowledge Distillation with Optimal Transport (PKDOT) method aims to enhance the performance of the student model by closing the gap between teacher and student representations. It relies on a cosine similarity matrix computed in the teacher and student space to capture the fine-grained structural information effectively. Entropy-regularized OT is then applied to minimize the distance between the teacher and the student. T-Net is also introduced to hallucinate the privileged modality features. Further, we select anchors based on dissimilarity to aid the OT in becoming more stable and introduce sparsity. Selecting anchor points instead of distilling from all samples is motivated by the sparsity notion in OT, where fewer samples contribute substantially. In contrast, if all samples are included in the distillation process, the model can be penalized based on less relevant samples for the class boundaries. Fig.~\ref{fig:pkdot_main} shows the architecture of the PKDOT method. The rest of this section provides additional details on the structural similarity matching and optimal transport in the PKDOT method.

\subsection{Structural Similarity Matching}
Let $\Theta$ be a matrix of dimension $b\times m$ obtained from the teacher network, where $b$ is the batch size and $m$ is the dimension of the feature vector, and $\Theta^\top$ be its transpose of dimension $m\times b$, and $\Phi$ be a matrix of dimension $b\times m$ obtained from the student network, and $\Phi^\top$ be its transpose of dimension $m\times b$.
{
\begin{equation} \label{calc_cst}
 C_\Theta = \Theta \cdot \Theta^\top
\text{,}\quad
 C_\Phi = \Phi \cdot \Phi^\top
\end{equation}
}
\noindent where $C_\Theta$ ($C_\Phi$) is a matrix of size $b\times b$ where each element $C_{ij}$ represents the dot product of $i^{th}$ row of $\Theta$ ($\Phi$) and $j^{th}$ column of $\Theta^\top$ ($\Phi^\top$)  and can be denoted as:
\begin{equation} \label{calc_elementct}
C_{ij}^\Theta = \sum_{k}^{} \Theta_{ik}.\Theta^{\top}_{kj}
\text{,}\quad
C_{ij}^\Phi = \sum_{k}^{} \Phi_{ik}.\Phi^{\top}_{kj}
\end{equation}

We normalize $C_\Theta$ and $C_\Phi$ to ensure that the similarity is not affected by the length. A normalization vector is calculated as follows:
\begin{equation} \label{norm_t}
N_{\Theta} = \sqrt{\sum_{k=1}^n (\Theta_{ik})^{2}}
\text{,}\quad
N_{\Theta}^ \top = \sqrt{\sum_{k=1}^n (\Theta^\top_{ik})^{2}}
\end{equation}
\noindent where $N_\Theta$ ($N_\Phi$)  is the normalization vector of size $b\times 1$, calculated by taking the $\ell^2$-norm of each row. $N_\Theta$ ($N_\Phi$) allows us to normalize the values so that the length of the vectors does not dominate the cosine similarity measure. $N_{\Theta}^\top$ ($N_{\Phi}^\top$) is also calculated using Eq.~\eqref{norm_t}, with the exception that it is of dimension $1\times b$.


$S_\Theta$ and $S_\Phi$ are the final similarity matrices of teacher and student networks respectively and are obtained by element-wise division of $C_\Theta$ with the dot product of $N{_\Theta}$ and $N_\Theta^ \top$ and $C_\Phi$ by the dot product of $N_\Phi$ and $N_{\Phi}^\top$ as follows: 
\begin{equation} \label{sim_mat}
S_\Theta = \frac {C_\Theta} {N_\Theta \cdot N_{\Theta}^\top}
 \text{,} \quad
S_\Phi = \frac {C_\Phi} {N_\Phi \cdot N_{\Phi}^\top }
\end{equation}






\subsection{Optimal Transport}

After the cosine similarity matrices have been computed for the given batch, the top-$k$ most dissimilar samples are selected based on the cosine similarity, and a new similarity matrix of dimension $b\times k$ is obtained. The same indexes are also selected in the student similarity matrix. The \textit{structural dark knowledge} is distilled from the teacher to the student using entropy-regularized OT, defined by:
\begin{equation}\label{eq:ot}
W_{p}(\mu ,\nu) = \int_{\chi \times \chi}^{} \mathcal{O}(S_{\Theta i},S_{\Phi i}) d\pi(S_{\Theta i},S_{\Phi i})) + \epsilon H(\pi)
\end{equation}
\noindent where $S_{\Theta i}$ and $S_{\Phi i}$ refer to each row in the teacher and student similarity matrices, respectively, semantically representing the local structures, and $\mu$ and $\nu$ are the marginal distributions. $\epsilon$ $>$ 0 is a coefficient and $H(\pi)$ is the entropic regularization defined as:
\begin{equation} \label{eq:entropic_reg}
    H(\pi) = log \left( \frac{d \pi}{d\mu \cdot d\nu} (S_{\Phi i},S_{\Theta i}) \right)
\end{equation}
Overall, ${W_p}$ represents an approximation of the Wasserstein distance between the teacher and student similarity matrices. This term is then included as an additional regularization to the task loss to align the intermediate representations.

The task loss, either concordance correlation coefficient (CCC) for regression (Eq.~\eqref{eq:task_loss_ccc}) or classification (Eq.~\eqref{eq:task_loss_ce}) loss, is calculated with the ground truth ($y$) as follows:  
\begin{equation} \label{eq:task_loss_ccc}
\mathcal{L}_{gt} =1 - \frac{2. \rho \cdot \sigma_{x} \cdot \sigma_{y}} {\sigma^{2}_{x} + \sigma^{2}_{y} + (\mu_{x}-\mu_{y})^{2}}
\end{equation}

\begin{equation} \label{eq:task_loss_ce}
\mathcal{L}_{gt} = -\sum_{c=1}^My_{o,c}\log(p_{o,c})
\end{equation}

The student network is then trained using: 
\begin{equation} \label{eq:total_loss}
\mathcal{L}_{total} = \mathcal{L}_{gt} + \lambda \cdot {W_{p}}
\end{equation}
\noindent where, $L_{GT}$ is the task loss, $W_p$ is the OT loss and $\lambda$ is the coefficient to weigh the importance of the OT loss.

\section{Experimental Methodology}

 \begin{figure*}[!t]
  \centering
   \includegraphics[width=0.85\linewidth]{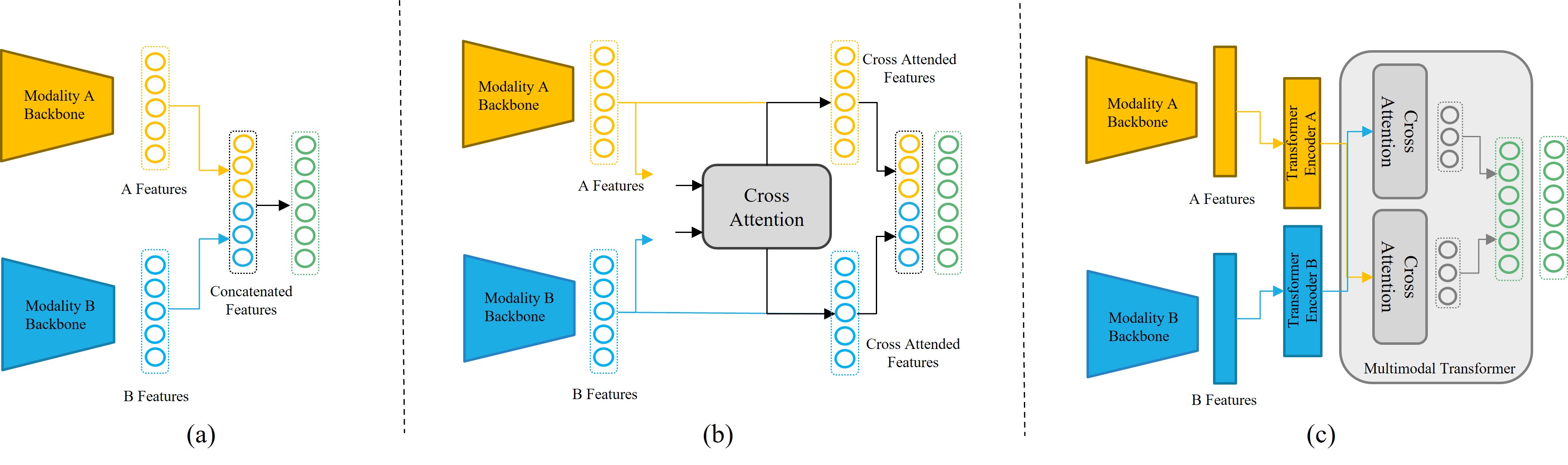}
   \caption{Illustration of different fusion mechanisms employed to obtain the joint teacher representation: (a) feature concatenation, (b) joint cross attention \cite{rajasekhar}, (c) multimodal transformer.}
   \label{fig:teacher_arch}
   \vspace{-10pt}
\end{figure*}

\subsection{Datasets and Evaluation Protocol}
\noindent \textbf{1) Biovid Heat Pain Database} is a popular database for pain estimation. The Part A of the dataset contains facial videos, ECG, EMG, and GSR. Part B of the dataset also contains facial EMG from the trapezius and corrugator muscles. The dataset is annotated for discrete labels for pain intensities, where BL refers to 'no pain' and PA1-PA4 refers to increasing pain intensities. The Biovid (A) dataset contains 8700 videos across 87 subjects, where each subject has 100 videos, corresponding to 20 videos per class. Some studies report results on 67 subjects and drop 20 subjects who do not exhibit any response to the pain. We use the entire dataset and report our results on the 87 subjects. The Biovid (B) dataset is recorded in a similar setting to Part A, with the addition of facial EMG, and it has a total number of 86 subjects, corresponding to 8600 videos. The dataset has no predefined train test split, so most studies validate using cross-validation. Following that, we perform 5-fold cross-validation.

\noindent \textbf{1) Affwild2 } is one of the largest datasets of in-the-wild videos gathered from YouTube \cite{affwild}. It contains 564 videos of varying length. The dataset has annotations for three affective computing problems: categorical expression recognition, action unit detection, and valence/arousal prediction. We validate the proposed method on the valence/arousal estimation problem. 
The dataset has training, validation, and test subsets with 351, 71, and 152 videos, respectively. This dataset is part of the ABAW challenge, so the test set annotations are not publicly available. Consequently, many studies report their results on the validation set.

\subsection{Implementation Details}

\subsubsection{\textit{Biovid dataset}}

\textbf{Teacher} For the Biovid (B) dataset, the multimodal teacher model is a multimodal transformer-based fusion model (Fig.~\ref{fig:teacher_arch}c) \cite{waligora2024joint}. The physiological backbone is a ResNet-18 that takes spectrograms of size 67$\times$127 obtained from EMG signals. The visual backbone is an R3D that takes cropped and aligned facial frames of size 112$\times$112 as input. The batch size for the visual backbone is 64, with a learning rate of 10$^{-3}$ 
optimized using the Adam optimizer. Both modalities output a feature vector of dimension 512. These feature vectors are fed to two separate transformer encoders and then to a multimodal transformer. K and V vectors are generated from one modality, and Q is generated from another. The output of the cross-attended vectors is gated using learnable weights. Finally, FC layers are added to generate the final prediction. The multimodal transformer is optimized with a learning rate of 10$^{-4}$ 
and the Adam optimizer with a batch size of 64. For the Biovid (A), we evaluate the proposed method with feature concatenation (Fig.~\ref{fig:teacher_arch}a), a relatively more straightforward fusion method. \textbf{Student:} The visual backbone and multimodal transformer model in the student network are the same as the teacher. The physiological backbone is dropped, and the T-Net is added to hallucinate the features of the privileged modality. The batch size for the student was set to 128, with a learning rate of 10$^{-4}$ 
and Adam optimizer and the value for $\lambda$  is 0.4.

\subsubsection{\textit{Affwild2 dataset}} \label{subsec:affwild}

\textbf{Teacher:} The joint cross-attention model (Fig.~\ref{fig:teacher_arch}b) proposed by Rajasekhar et al.~\cite{praveen-tbbi} is used. We use the cropped-aligned facial images that are provided with the dataset. A 3D CNN is used to extract features from the visual modality. The batch size for the visual modality is 8, and the learning rate used is 10$^{-3}$ 
The audio modality is divided into multiple short segments corresponding to 256 frames in the visual modality. Spectrograms of resolution 64$\times$107 are obtained using DFT. These spectrograms are fed to a ResNet-18 trained from scratch to extract the audio features. The batch size is the same as the visual modality with a learning rate of 10$^{-3}$. For the fusion of the two modalities, the concatenated vector of size 1024 is fed to the cross-attention module, which is optimized using the Adam optimizer with a learning rate of 10$^{-3}$ 
and a batch size of 64. \textbf{Student:} The student network follows roughly the same implementation method as the teacher network, except there is no backbone to process the privileged modality. Instead, it is replaced with a T-Net, a simple encoder-decoder trained during the teacher training phase. In the student training phase, the T-Net is frozen and is used to hallucinate the features of the privileged modality. The batch size is set to 128 for the student model, and the value for $\lambda$ is 0.4.

\section{Results and Discussion}

\subsection{Comparison with the State-of-the-Art}

Table \ref{tab:summ-mod} provides an overview of the experimental settings, including different modalities, fusion methods, and the distillation setting. Table \ref{tab:pkd-sota} compares the proposed method with SOTA-privileged distillation methods. We compare PKDOT with cosine similarity \cite{pkd-aslam}, MSE, and KL \cite{pkd-makantasis}. It can be observed from the table that PKDOT outperforms point-to-point distillation methods like KL-divergence, MSE, and cosine similarity. The improvement for the BioVid dataset is considerable, partly because the visual modality is weak in the pain estimation problem and because multiple subjects show little to no variation for pain and no-pain videos. Transporting the PI, i.e., EMG, from the multimodal network to the visual network greatly enhances the performance. The improvement in the Affwild2 dataset is marginal because the PI, i.e., audio modality, is weak on its own but can still improve the visual-only network. This also suggests that the proposed network can be used in stronger-enhancing-weaker (SEW) and weaker-enhancing-stronger (WES) scenarios.

\begin{table}[]
\begin{tabular}{c|c|l|c|c}
\Xhline{1.55pt}
\textbf{Dataset}                                                                & \textbf{Modalities} & \textbf{Property}                                                      & \textbf{Fusion}                                                                            & \textbf{Distilled to}                                                             \\ \Xhline{1.55pt}
\multirow{2}{*}{Affwild}                                               & Visual     & \begin{tabular}[l]{@{}l@{}}-Strong\\ -Prevalent\end{tabular}  & \multirow{2}{*}{\begin{tabular}[c]{@{}c@{}}Joint Cross \\ Attention\end{tabular}} & \multirow{2}{*}{\begin{tabular}[l]{@{}c@{}}Visual\\ (WES)\end{tabular}}  \\ \cline{2-3}
                                                                       & Audio      & \begin{tabular}[l]{@{}l@{}}-Weak\\ -Privileged\end{tabular}   &                                                                                   &                                                                          \\ \hline
\multirow{2}{*}{\begin{tabular}[l]{@{}l@{}}Biovid\\  (A)\end{tabular}} & Visual     & \begin{tabular}[l]{@{}l@{}}-Weak\\ -Prevalent\end{tabular}    & \multirow{2}{*}{\begin{tabular}[c]{@{}c@{}}Feature\\ Concatenation\end{tabular}}  & \multirow{2}{*}{\begin{tabular}[l]{@{}l@{}}Visual\\ (SEW)\end{tabular}}  \\ \cline{2-3}
                                                                       & EDA        & \begin{tabular}[l]{@{}l@{}}-Strong\\ -Privileged\end{tabular} &                                                                                   &                                                                          \\ \hline
\multirow{2}{*}{\begin{tabular}[l]{@{}l@{}}Biovid\\ (B)\end{tabular}}  & Visual     & \begin{tabular}[l]{@{}l@{}}-Strong\\ -Prevalent\end{tabular}  & \multirow{2}{*}{\begin{tabular}[l]{@{}l@{}}Multimodal\\ Transformer\end{tabular}} & \multirow{2}{*}{\begin{tabular}[l]{@{}l@{}}Visual \\ (WES)\end{tabular}} \\ \cline{2-3}
                                                                       & EMG        & \begin{tabular}[l]{@{}l@{}}-Weak\\ -Privileged\end{tabular}   &                                                                                   &                                                                          \\ \Xhline{1.55pt}
\end{tabular}
\caption{Summary of diverse modalities, their properties, fusion mechanisms, and distillation settings: Stronger Enhancing Weaker (SEW) and Weaker Enhancing Stronger (WES).}
\label{tab:summ-mod}
\end{table}

 \begin{figure*}[t]
  \centering
   \includegraphics[width=1\linewidth]{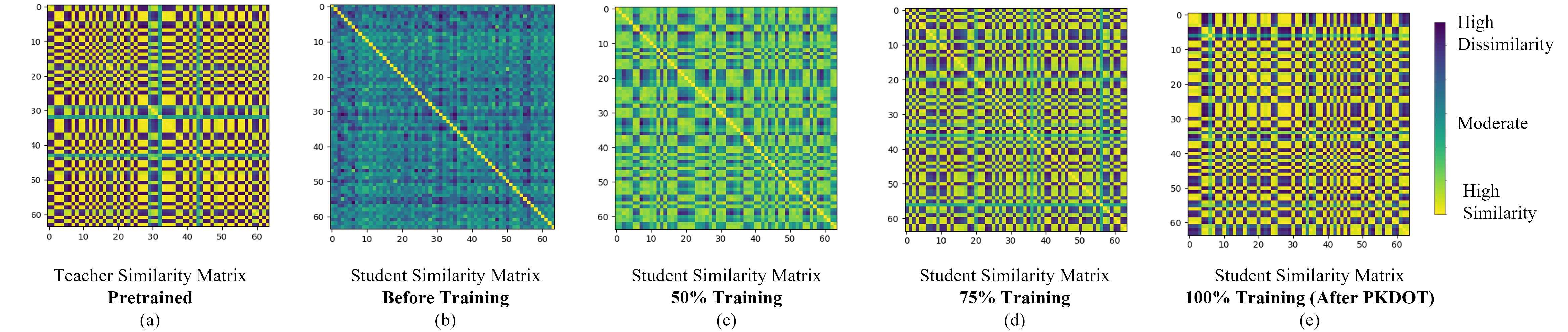}
   \caption{Evolution of the similarity matrix over training epochs.}
   \label{fig:simmat_5}
   \vspace{-5pt}
\end{figure*}

Fig.~\ref{fig:simmat_5} shows the evolution of the similarity matrix with training. Fig.~\ref{fig:simmat_5}a is the similarity matrix for the pretrained teacher. It can be seen from the image that the samples from the same class have high similarity, and the samples from the other class have high dissimilarity. Fig.~\ref{fig:simmat_5}b is the similarity matrix for the student before training, which shows a moderate similarity with other samples, and the diagonal represents the high similarity between the sample and itself. As the training progresses, we see the student similarity matrix become closer and closer to that of the teacher. 
\begin{table}[t]
\centering 
\setlength{\tabcolsep}{0.45em}
\begin{tabular}{cc|cc}
\Xhline{1.55pt}
\multicolumn{2}{c|}{\textbf{Method}}                                                                                                     & \multicolumn{2}{c}{\textbf{Performance}} \\  
\multicolumn{2}{c|}{}                                                                                                                    & Biovid (B)             & Affwild             \\ \Xhline{1.55pt}
Aslam et al. \cite{pkd-aslam}                           &  \begin{tabular}[c]{@{}c@{}}\textbf{Cosine Similarity} \\ \textit{(IEEE CVPRw '23)}\end{tabular}      &      75.40              &       \begin{tabular}[c]{@{}l@{}}V: 0.37 \\ A: 0.53 \end{tabular} \\  \hline
Makantasis et al. \cite{pkd-makantasis}                 & \begin{tabular}[c]{@{}c@{}}\textbf{MSE} \\  \textit{(IEEE TAC '23)}\end{tabular}                     &      N/A           &         \begin{tabular}[c]{@{}l@{}}V: 0.39 \\ A: 0.53\end{tabular} \\  \hline
Makantasis et al. \cite{pkd-makantasis} & \begin{tabular}[c]{@{}c@{}}\textbf{KL} \\ \textit{(IEEE TAC '23)}\end{tabular}                      &      75.80              &     N/A \\               \Xhline{1.55pt}
PKDOT (Ours)                                              & \begin{tabular}[c]{@{}c@{}}Optimal Transport\\  Structural KD\end{tabular} &        \textbf{78.76}            &      \begin{tabular}[c]{@{}l@{}}\textbf{V: 0.43} \\ \textbf{A: 0.56} \end{tabular} \\                \Xhline{1.55pt}
\multicolumn{4}{l}{N/A: Not applicable to the problem.}
\end{tabular}
\caption{Performance of the proposed and SOTA privileged KD methods on Biovid and Affwild2 datasets.}
\label{tab:pkd-sota}
\vspace{-10pt}
\end{table}

\begin{table}[]
\centering
\setlength{\tabcolsep}{0.45em}
\begin{tabular}{c|c|c|c}
\Xhline{1.55pt}
\textbf{Method}                                                                   & \textbf{Visual Network} & \textbf{Valence} & \textbf{Arousal} \\ \Xhline{1.55pt}
Baseline  \cite{aff-baseline} \textit{CVPRw'20}                                                                      & ResNet-50               & 0.31             & 0.17             \\ \hline
Zhang et al. \cite{Zhang-affwild} \textit{CVPRw'20}                                                                    & SENet-50                & 0.28             & 0.34             \\ \hline
He et al.  \cite{he-affwild} \textit{CVPRw'21}                                                                        & MobileNet               & 0.28             & 0.44             \\ \hline
Nguyen et al.  \cite{nguyen-affwild} \textit{CVPRw'22}                                                                    & RegNet + GRU            & 0.43             & 0.57             \\ \hline
Geesung et al. \cite{geesung} \textit{CVPRw'21}                                                                    & ResNeXt + SENet         & 0.51             & 0.48             \\ \Xhline{1.55pt}
Visual only (Lower bound) & \multirow{3}{*}{I3D}    & 0.41             & 0.51             \\ \cline{1-1} \cline{3-4} 
\textbf{Student - PKDOT (Ours)}                                                             &                         & \textbf{0.43}          & \textbf{0.56}             \\ \cline{1-1} \cline{3-4} 
\begin{tabular}[c]{@{}c@{}}Multimodal (Visual + Audio)\\ (Upper bound)\end{tabular}                 &                         & 0.67             & 0.59             \\\Xhline{1.55pt}
\end{tabular}
\caption{Performance of the proposed with SOTA visual-only methods on the Affwild2 validation set.}
\label{tab:aff-sota}
\vspace{-5pt}
\end{table}

Table \ref{tab:aff-sota} compares the student model with visual-only SOTA on the Affwild2 dataset. The multimodal upper bound in the Affwild2 dataset for valence and arousal is 0.67 and 0.59, respectively. The visual-only network is the lower bound with 0.41 valence and 0.51 arousal value. The PKDOT student model shows an improvement of absolute 2\% for the valence dimension and 3\% for the arousal dimension. Table \ref{tab:biovida-sota} compares the PKDOT student model with visual-only methods on the Biovid (A) dataset. Some studies report results on no-pain (BL1) and pain (PA4), and some have included multi-class (MC), i.e., a 5-class problem. We present and compare both settings. It can be seen from the table that the proposed system gains an absolute 2.4\% over the visual-only lower bound.

\begin{table}[t]
\centering
\setlength{\tabcolsep}{0.45em}
\begin{tabular}{c|c|cc}
\Xhline{1.55pt}
\multirow{2}{*}{\textbf{Method}}                                                         & \multirow{2}{*}{\textbf{Visual Network}} & \multicolumn{2}{c}{\textbf{Accuracy}}             \\ \cline{3-4} 
                                                                                &                               & \multicolumn{1}{c|}{BL1 vs. PA4} & MC    \\ \Xhline{1.55pt}
Zhi et al.   \cite{zhi-biovid} \textit{ITAIC'19}                                                                  & Sparse LSTM                   & \multicolumn{1}{c|}{61.70}        & 29.70  \\ \hline
Morabit et al.  \cite{morabit-biovida} \textit{ISIVC'22}                                                               & ViT                           & \multicolumn{1}{c|}{72.11}       & NR     \\ \hline
Werner et al. \cite{werner-biovida} \textit{ACIIW'17}                                                                 & FAD + RF                      & \multicolumn{1}{c|}{72.40}        & 30.80  \\ \hline
Patania et al. \cite{patania-biovid_a} \textit{SIGAPP'22}                                                                & GNN                           & \multicolumn{1}{c|}{73.20}        & NR     \\ \hline
Gkikas et al. \cite{gkikas-biovida} \textit{EMBC'23}                                                              & ViT                           & \multicolumn{1}{c|}{73.28}       & 31.52 \\ \hline
 Dragomir et al. \cite{dragomir-biovidA} \textit{EHB'20}                                                               & ResNet-18                     & \multicolumn{1}{c|}{NR}           & 36.60  \\ \Xhline{1.55pt}
Visual only (Lower bound) & \multirow{3}{*}{R3D}          & \multicolumn{1}{c|}{72.10}        & 30.38  \\ \cline{1-1} \cline{3-4} 
\textbf{Student - PKDOT (ours) }                                                       &                               & \multicolumn{1}{c|}{\textbf{74.55}}         & \textbf{33.65}  \\ \cline{1-1} \cline{3-4} 
\begin{tabular}[c]{@{}c@{}}Multimodal (Visual + EDA)\\ (Upper bound)\end{tabular}            &                               & \multicolumn{1}{c|}{83.50}        & 36.40   \\ \Xhline{1.55pt}
\multicolumn{4}{l}{\scriptsize NR: Not Reported.}
\end{tabular}
\caption{Performance of the proposed and SOTA visual-only methods on the Biovid (A) dataset.}
\label{tab:biovida-sota}
\vspace{-10pt}
\end{table}

\begin{table}[t]
\centering
\begin{tabular}{c|c|c}
\Xhline{1.55pt}
\textbf{Method}                                                                         & \textbf{Visual Network }      & \textbf{Accuracy} \\ \Xhline{1.55pt}
Kachele et al. \cite{Kchele2015BioVid_mm} \textit{IWMCS'15}                                                                & Handcrafted + RF     & 72.70    \\ \Xhline{1.55pt}

Visual Only (Lower bound) & \multirow{3}{*}{R3D} & 74.10    \\ \cline{1-1} \cline{3-3} 
\textbf{Student - PKDOT (Ours)}                                                          &                      & \textbf{78.76}    \\ \cline{1-1} \cline{3-3} 
\begin{tabular}[c]{@{}c@{}}Multimodal (Visual + EMG)\\ (Upper bound)\end{tabular}
             &                      & 81.30   \\ \Xhline{1.55pt}
\end{tabular}
\caption{Performance of the proposed and SOTA visual-only methods on the Biovid (B) dataset.}
\label{tab:biovid-sota}
\vspace{-15pt}
\end{table}

Table \ref{tab:biovid-sota} compares the proposed method with methods in the literature for the Biovid (B) dataset. It can be observed from the table that the proposed PKDOT student model has gained an absolute 4.6\% over the visual-only model. 

Note that other methods in the literature may be tailored to the problem, with sophisticated architecture changes or extensive pretraining to achieve higher performance. This work aims not to push the SOTA on these problems but to exhibit that PKDOT can enhance performance by using simple models and only the visual modality at inference. 

\subsection{Ablations}
\subsubsection{Batch Size and Anchor Selection}
\begin{figure}[!h]
 \centering
  \includegraphics[width=0.78\linewidth]{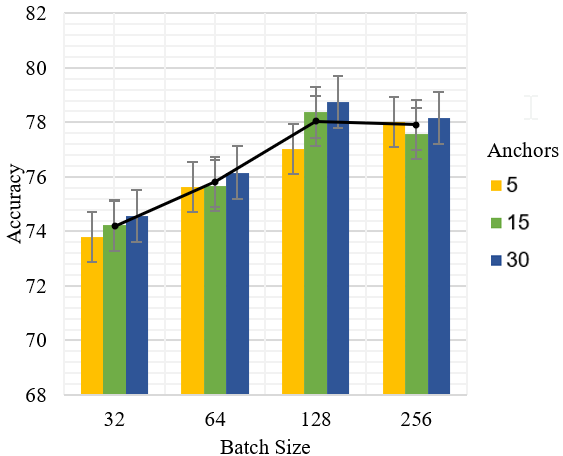}
  \caption{Evolution of the performance of the proposed network with batch size and number of anchors variation on the Biovid (B) dataset.}
  \label{fig:ab-bs-chart}
    \vspace{-10pt}
\end{figure}

Batch size is an essential parameter in the proposed mechanism since it dictates how many samples are included in the similarity matrix at each optimization step. We run PKDOT with batch sizes from 32 to 256; for each experiment, we select 5, 15, and 30 anchors. Fig.~\ref{fig:ab-bs-chart} shows the evolution of results according to batch size and number of anchors. Our experiments show that increasing the batch size shows an improvement in the distillation efficiency. The best results are obtained with the batch size of 128 and 30 anchors. 


\subsubsection{OT Regularization}
Entropy-regularized OT has a regularization coefficient that is the trade-off between minimizing transportation costs and maintaining uniformity.


We run experiments with different values of $\epsilon$ in Eq.~\ref{eq:entropic_reg}. 

\begin{figure}[!h]
 \centering
  \includegraphics[width=0.91\linewidth]{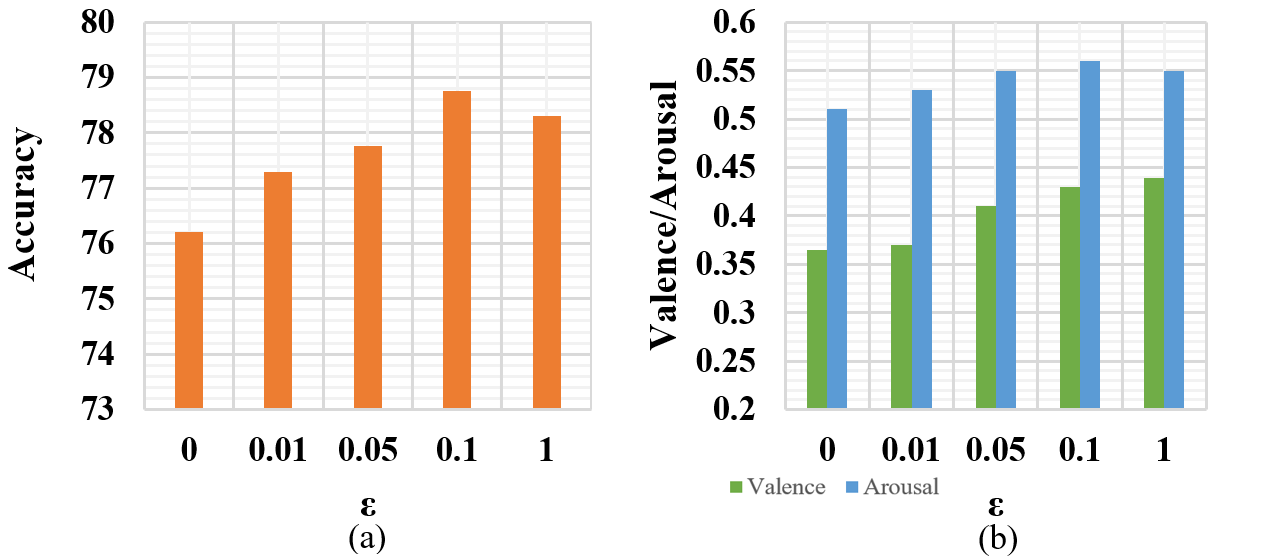}
  \caption{The impact of the PKDOT regularization parameter $\epsilon$ on (a) accuracy for the Biovid dataset and (b) valence/arousal for the Affwild2 dataset.}
  \label{fig:ab-reg-chart}
  \vspace{-10pt}
\end{figure}

Fig.~\ref{fig:ab-reg-chart} shows the impact of the regularization parameter on the PKDOT student performance. We run experiments with different values of $\epsilon$. Larger values of $\epsilon$ emphasize the regularization term, i.e., it encourages the transport plan to be more sparse, where key samples contribute substantially. Values closer to zero put more emphasis on minimizing the transportation cost. In our experiments, $\epsilon$=0.1 gives the best trade-off between the transport cost and uniformity.

 The proposed method is validated with different fusion architectures because the method used to obtain the multimodal teacher representation also has a bearing on distillation efficiency. We test with both, simple and complex fusion architectures. WES is when the privileged modality is weaker and the prevalent modality is stronger. SEW is when the stronger modality is the privileged one, and the student model only inputs the weaker modality. The experiments show that the proposed method works better in WES settings. However, there is more margin to enhance the performance when the prevalent modality is weaker since there is a significant performance gap between the student and teacher performance. It is still more burdensome for the student model to mimic the teacher because the stronger modality is unavailable to the student. In WES settings, the absent modality is weaker in comparison, and the student still has access to the stronger modality that the teacher model had used to learn the discriminative information.

\section{Conclusions}

This paper presents a novel OT-based structural KD mechanism for transferring privileged information. Most privileged KD methods employ point-to-point matching, which cannot capture the local structures in the teacher space. The proposed PKDOT method solves this by capturing the local structures and applying entropy-regularized OT to translate the student representation closer to the teacher representation. To further support the OT solution for effective distillation, we select top-k anchors and only calculate the OT between them. The proposed PKDOT method is validated on two main MER problems: the pain classification and the valence/arousal estimation regression problem. To further prove the generalization capability of our proposed PKDOT method, we employ three different fusion methods and a variety of modalities to form the teacher representations, showing that the proposed method is model-agnostic and modality-agnostic. The proposed method can significantly improve student performance over the SOTA methods using privilege KD in both SEW and WES settings. 

\noindent \textbf{\textit{Limitations:}} The proposed method only allows for knowledge transfer after the modalities have been fused in the teacher network. 
Further, the model assumes that the top-k most dissimilar anchors are the most effective and contribute the most to the distillation process. Identifying the most effective anchor points through learning is a future research direction.

\section{ACKNOWLEDGMENTS}
This work was supported in part by the Fonds de recherche du Québec – Santé (FRQS), the Natural Sciences and Engineering Research Council of Canada (NSERC), Canada Foundation for Innovation (CFI), and the Digital Research Alliance of Canada.


{\small
\bibliographystyle{unsrt}
\bibliography{egbib}
}
\balance
\end{document}